\documentclass[conference]{IEEEtran}
\IEEEoverridecommandlockouts
\usepackage{cite}
\usepackage{amsmath,amssymb,amsfonts}
\usepackage{graphicx}
\usepackage{subcaption}
\usepackage{textcomp}
\usepackage{enumitem}
\usepackage{xcolor}
\usepackage{algorithm}
\usepackage{algpseudocode}
\usepackage{xcolor}
\usepackage{booktabs} % add to preamble
\usepackage[colorlinks]{hyperref}
\hypersetup{
    citecolor=blue, linkcolor=blue, urlcolor=blue
    }
\usepackage{xcolor}
\begin{document}

\title{Z-Pruner: Post-Training Pruning of Large Language Models for Efficiency without Retraining}

\author{
\IEEEauthorblockN{
Samiul Basir Bhuiyan\textsuperscript{1},
Md. Sazzad Hossain Adib\textsuperscript{1},
Mohammed Aman Bhuiyan\textsuperscript{1},
Muhammad Rafsan Kabir\textsuperscript{1},\\
Moshiur Farazi\textsuperscript{2},
Shafin Rahman\textsuperscript{1},
Nabeel Mohammed\textsuperscript{1}}

\IEEEauthorblockA{\textsuperscript{1}Department of Electrical and Computer Engineering, North South University, Dhaka, 1229, Bangladesh}

\IEEEauthorblockA{\textsuperscript{2}Data Science and AI, University of Doha for Science and Technology, Doha, Qatar}

\textsuperscript{1}\{samiul.bhuiyan, sazzad.adib, mohammed.aman, muhammad.kabir, shafin.rahman, nabeel.mohammed\}@northsouth.edu \\
\textsuperscript{2}moshiur.farazi@udst.edu.qa
}

% \author{
% \IEEEauthorblockN{Anonymous Author(s)}
% }
\maketitle

\begin{abstract}
Large language models (LLMs) have rapidly advanced in recent years, achieving remarkable performance across a wide range of natural language processing tasks. However, this progress has come at the cost of increasingly large model sizes, which pose significant challenges for deployment, scalability, and energy efficiency. To address these limitations, post-training pruning has emerged as a promising approach for reducing model size and inference latency without the need for retraining. Despite these advantages, many existing pruning methods result in substantial performance degradation or require computationally expensive fine-tuning. In this work, we introduce Z-Pruner, a novel post-training pruning method designed to induce sparsity in pretrained LLMs without any retraining. Unlike conventional approaches, Z-Pruner leverages both weight update magnitudes and activation patterns to identify and eliminate redundant parameters more effectively. Our method is model-agnostic, efficient, and easy to implement. We evaluate Z-Pruner using multiple widely-used LLM architectures, including LLaMA-2, LLaMA-3, and OPT, across a diverse set of standard language benchmarks. Experimental results demonstrate that Z-Pruner surpasses state-of-the-art pruning methods that require intensive weight updates. Specifically, Z-Pruner achieves the lowest perplexity scores and the highest overall average score for zero-shot accuracy. We have made the corresponding codes publicly available at \href{https://github.com/sazzadadib/Z-Pruner}{https://github.com/sazzadadib/Z-Pruner}.
\end{abstract}

\begin{IEEEkeywords}
Large Language Model, Pruning, Perplexity
\end{IEEEkeywords}

\section{Introduction}
Large Language Models (LLMs) have transformed natural language processing (NLP), demonstrating remarkable capabilities across diverse and challenging tasks \cite{kabir2025legalrag, bommarito2022gpttakesbarexam, wei2022emergentabilitieslargelanguage, rahman2019transductive, bubeck2023sparksartificialgeneralintelligence}. Their ability to perform complex reasoning, summarization, translation, and question answering has made them indispensable in both academic research and industrial applications. However, the tremendous performance of LLMs comes at a steep price: their massive scale, often spanning billions of parameters, demands enormous computational resources, including high-end hardware accelerators and extensive memory footprints. Addressing this challenge is critical to democratize the power of LLMs and enable their deployment on resource-constrained devices.

Several works have contributed to addressing this issue by exploring model compression techniques. Among these, model quantization \cite{dettmers2022llmint88bitmatrixmultiplication, frantar2023gptqaccurateposttrainingquantization, cheraghian2021synthesized, xiao2024smoothquantaccurateefficientposttraining, ahmadian2023intriguingpropertiesquantizationscale} reduces precision to lower-bit representations, achieving remarkable efficiency gains, yet often struggles to fully eliminate redundancy in weights. Quantization reduces the precision of weights and activations, but this coarse representation can introduce rounding errors that negatively affect model accuracy. Another widely used approach is network pruning \cite{hossain2025colt, NIPS1989_6c9882bb, NIPS2015_ae0eb3ee}, a complementary technique that removes weights with the smallest absolute values by setting them to zero. However, this approach can overlook the collective contribution of these weights to the model's performance, potentially causing unexpected accuracy drops. Traditional pruning algorithms require model retraining \cite{liu2019rethinkingvaluenetworkpruning, blalock2020stateneuralnetworkpruning}, training from random initializations \cite{louizos2018learningsparseneuralnetworks}, or costly iterative rewinding and fine-tuning procedures \cite{frankle2019lotterytickethypothesisfinding, renda2020comparingrewindingfinetuningneural}.

\begin{figure}[t]
\centering
\includegraphics[width=0.5\textwidth]{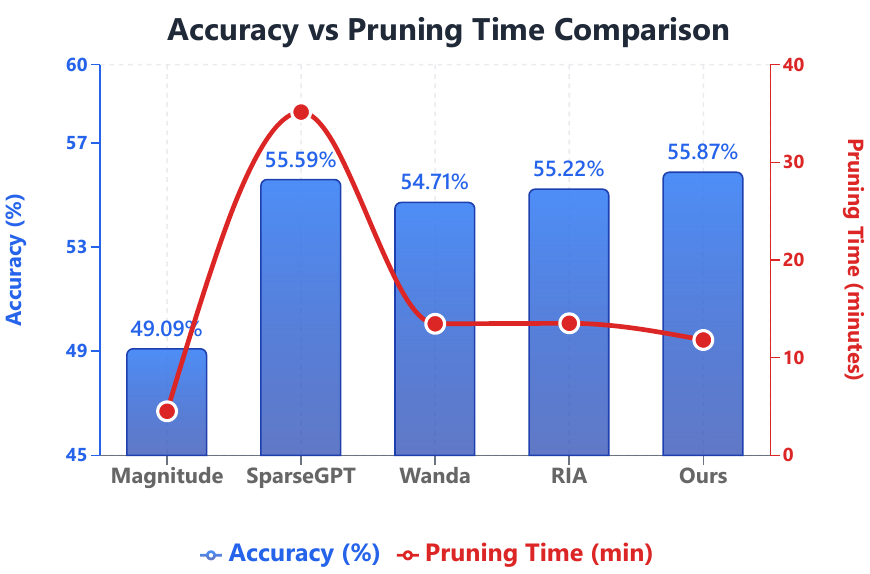} % Adjust the width as needed
\caption{Comparison of pruning methods in terms of zero-shot average accuracy and pruning time. Our method achieves the highest accuracy with low pruning time.}
\label{fig1}
\end{figure}

These processes are highly resource-intensive, making them impractical for billion-scale LLMs. In general, sparsity can be induced through three routes: \emph{(a)} sparse training, which integrates sparsity into the training process itself \cite{lee2019snipsingleshotnetworkpruning, evci2021rigginglotterymakingtickets, sanh2020movementpruningadaptivesparsity}; \emph{(b)} pruning-aware training, which adapts the learning process to accommodate pruning constraints \cite{han2015learningweightsconnectionsefficient, liu2022sparsetrainingboostingpruning}; and \emph{(c)} post-training pruning (PTP), which compresses an already trained model without re-training \cite{frantar2023sparsegptmassivelanguagemodels, sun2024simpleeffectivepruningapproach}. Nevertheless, sparse and pruning-aware training methods are typically impractical for LLMs due to their requirement for multiple training passes and high memory demands, while even state-of-the-art PTP techniques like SparseGPT still involve expensive weight update steps that scale poorly with model size.

To address these challenges and limitations, we develop a fast and effective post-training pruning method for large language models that avoids retraining or costly update procedures. Specifically, we introduce Z-Pruner, a statistically principled post-training pruning technique that leverages weight normalization and activation-aware scaling to identify redundant parameters, enabling adaptive sparsification of LLMs in a single pass. Our evaluations across multiple benchmark datasets demonstrate that the proposed post-training pruning algorithm achieves lower perplexity scores and higher average zero-shot accuracy with lower pruning time compared to existing state-of-the-art pruning algorithms, as presented in Figure \ref{fig1}.

The major contributions are summarized as follows:
\begin{itemize}
\item We propose Z-score pruning, a statistically grounded post-training criterion that normalizes weights and amplifies outliers to identify redundant parameters without retraining.
\item We combine statistical weight importance with activation-aware scaling, introducing two activation functions: one specialized for OPT models and another tuned for LLaMA models, achieving architecture-aware pruning.
\item Z-Pruner achieves state-of-the-art perplexity and zero-shot accuracy under 50\% sparsity on multiple benchmarks using several different LLMs, including OPT 6.7B, LLaMA-2 7B/13B, and LLaMA-3.1 8B, while providing faster pruning than prior PTP methods.
\end{itemize}

\section{Related Work}

\subsection{Pruning of Large Language Models}
Pruning is a classic model compression technique used to reduce the size and computational cost of neural networks. It is especially important for Large Language Models (LLMs), which are often resource-intensive. Pruning techniques fall into two categories: unstructured pruning removes individual weights, producing sparse matrices that save storage but require specialized hardware for speedups \cite{NIPS1989_6c9882bb}; structured pruning removes larger components like attention heads, neurons, or layers, making models faster on standard hardware \cite{molchanov2017pruning}. Foundational work introduced pruning combined with quantization and Huffman coding to compress deep networks with minimal accuracy loss \cite{han2016deepcompressioncompressingdeep}. Structured pruning for LLMs has been advanced through low-rank matrix factorization \cite{Wang_2020}, which adaptively removes components during training to improve efficiency in tasks such as BERT fine-tuning. Recent work has proposed LLM-specific approaches, notably LLM-Pruner \cite{ma2023llmprunerstructuralpruninglarge}, a task-agnostic structured pruning framework that exploits gradient information to identify and eliminate non-essential components while preserving model generality. Additional techniques, such as movement pruning \cite{sanh2020movementpruningadaptivesparsity}, Taylor expansion–based saliency estimation \cite{molchanov2019importanceestimationneuralnetwork}, and sparsity-aware training \cite{yuan2021mestaccuratefastmemoryeconomic}, also aim to optimize the trade-off between computational efficiency and model accuracy. Our method advances unstructured LLM pruning by explicitly updating weight importance through a Z-score–based scoring mechanism, then pruning accordingly for improved performance.

\subsection{Post-Training Pruning}
Post-training pruning eliminates weights from a pre-trained model without additional training, a highly desirable trait given the cost of fine-tuning billion-scale LLMs. This approach leverages the redundancy in LLMs to reduce size while preserving functionality. A notable method is Wanda \cite{sun2024simpleeffectivepruningapproach} (Weights and Activations Pruning), which selects weights based on their magnitude scaled by activations. It introduces sparsity without retraining and performs competitively on benchmarks using models like LLaMA\cite{touvron2023llamaopenefficientfoundation}. Wanda's simplicity and strong performance have made it a benchmark in this space. Similarly, Plug-and-Play pruning uses Relative Importance and Activations (RIA) and Channel Permutation to enhance sparsity under N:M constraints \cite{zhang2024plugandplay}. Our method fits into the category of post-training pruning frameworks, using two activation functions across different families of LLM models (e.g., LLaMA, OPT), significantly improving the performance of large language models.

\subsection{Pruning Without Retraining}
Pruning without retraining is a subcategory of post-training pruning where the pruned model is used immediately, with no fine-tuning. This is challenging because the pruning must be precise enough to retain model performance. Wanda is a leading example, achieving strong results without any retraining by focusing on activation-informed weight selection \cite{sun2024simpleeffectivepruningapproach}. LLM-Pruner also supports this approach: its pruning step requires no retraining, though it allows optional lightweight tuning to regain performance \cite{ma2023llmprunerstructuralpruninglarge}. These methods show that with carefully designed pruning criteria, it is possible to deploy compressed LLMs efficiently even in resource-constrained environments. Our approach does not require any additional retraining or fine-tuning after pruning, ensuring the model retains performance while reducing parameter counts.

\section{Methodology}

In this section, we present Z-Pruner, a post-training pruning framework designed to efficiently compress large language models (LLMs) without any retraining. Z-Pruner combines a statistically principled weight-importance criterion with lightweight activation scaling to achieve sparsity while preserving performance. The proposed methodology can be divided into three key components: weight normalization, z-score–based importance scoring, and activation scaling.

\begin{figure*}[t]
\centering
\includegraphics[width=1\textwidth]{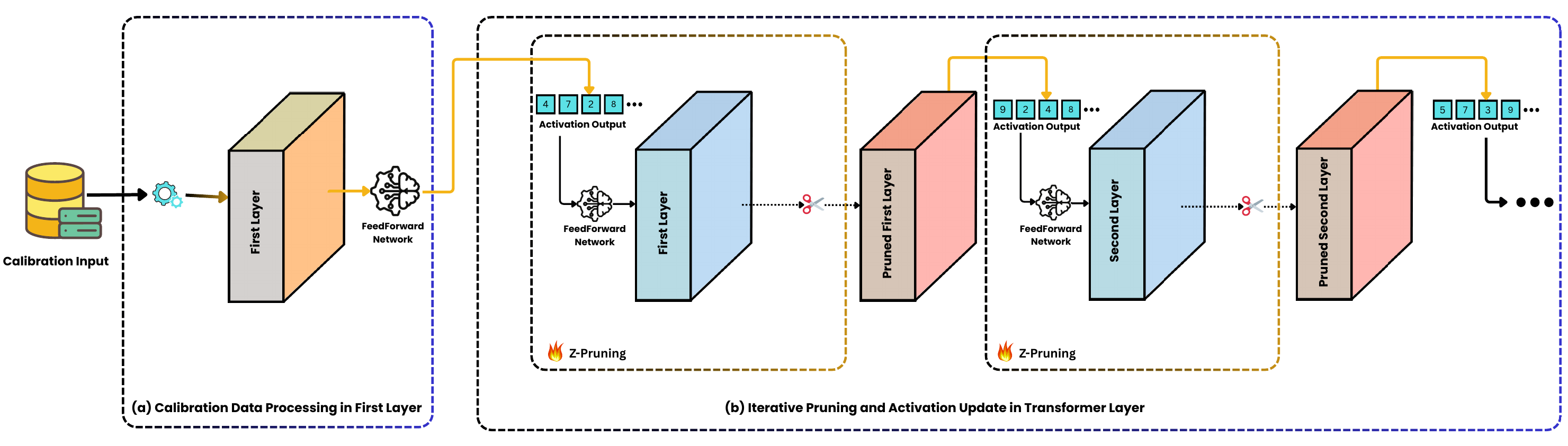}
\caption{\textbf{(a)} Illustration of calibration data being processed by the first layer of a transformer model to generate output values. \textbf{(b)} Illustration of the iterative process where input activations are processed by a transformer layer, followed by weight pruning, generating new activations for subsequent pruning iterations.}
\label{z-matrix}
\end{figure*}

\subsection{Problem Formulation}
Suppose a calibration dataset $\mathcal{D}_{\text{calib}}=\{x_i\}_{i=1}^n$, where $x_i \in \mathbb{R}^{s \times d}$ represents tokenized input sequences of length $s = 2048$ and hidden dimension $d$, and $n = 128$ is the number of calibration samples (e.g., from C4). For evaluation, we use a test dataset $\mathcal{D}_{\text{test}} = \{x_j, y_j\}_{j=1}^m$, where $x_j \in \mathbb{R}^{s \times d}$ are input sequences and $y_j \in \mathbb{R}^s$ are target tokens for next-token prediction, with $m$ samples from WikiText2. The dense model $\mathcal{M}$ without compression produces predictions $\mathcal{M}(x_i) = \hat{y}_i$, but incurs high memory and computational costs due to billions of parameters.

In this study, we propose a compression framework using the pruning method, Z-Pruner, to reduce model weights by a sparsity ratio $\rho = 0.5$ (adjustable), producing a sparse model $\mathcal{M}_{\text{pruned}}$ with unstructured sparsity. Given a target sparsity ratio $\rho$, we aim to construct a binary mask $M_{\text{mask}} \in \{0,1\}^{m \times n}$ such that the pruned weight matrix is given by $W' = W \odot M_{\text{mask}}$, where $\odot$ denotes element-wise multiplication. 
The mask $M_{\text{mask}}$ is determined by selective weights based on the importance scores in $M$ based on pruning mode. For global pruning, the matrix $M$ is flattened and sorted to identify the top $(1-\rho)\cdot mn$ where $mn$ is flatten matrix values across all entries or per-output-neuron pruning, each row of $M$ is independently sorted to select the top $(1-\rho)\cdot m$ values within that row, ensuring each output neuron retains its most important connections.

The proposed post-pruning approach ensures that:
\begin{align}
\mathcal{M}_{\text{pruned}}(x_i) &= \hat{u}_i \quad \text{where } \hat{u}_i \approx \hat{y}_i
\end{align}
The pruned model $\mathcal{M}_{\text{pruned}}$ minimizes memory footprint while preserving performance, with $\hat{u}_i \approx \hat{y}_i$ indicating that the sparse model produces approximately similar predictions despite slight performance degradation due to weight removal. The overall methodology is presented in Figure \ref{z-matrix}.

\begin{algorithm}
\caption{Z-Pruner}
\label{algo1}
\textbf{Inputs:} weight matrix $W \in \mathbb{R}^{m \times n}$ \\
\textbf{Output:} updated weight matrix $W'$
\begin{algorithmic}[1]
\State $\hat{W}^{(r)}_{ij} \gets W_{ij}/\|W_{i,:}\|_2$ 
\State $\hat{W}^{(c)}_{ij} \gets W_{ij}/\|W_{:,j}\|_2$ 
\State $\mu_r, \sigma_r \gets \operatorname{mean}(\hat{W}^{(r)}), \operatorname{std}(\hat{W}^{(r)})$
\State $\mu_c, \sigma_c \gets \operatorname{mean}(\hat{W}^{(c)}), \operatorname{std}(\hat{W}^{(c)})$
\State $D^{(r)}_{ij} \gets (\hat{W}^{(r)}_{ij} - \mu_r)/\sigma_r$
\State $D^{(c)}_{ij} \gets (\hat{W}^{(c)}_{ij} - \mu_c)/\sigma_c$
\State $I^{(r)}_{ij} \gets |D^{(r)}_{ij}|^3$
\State $I^{(c)}_{ij} \gets |D^{(c)}_{ij}|^3$
\State $s \gets |W_{ij}| < 0.1 \cdot \operatorname{mean}(|W|)$
\State $\alpha \gets 0.7 \cdot (1 - 0.3 \cdot s)$
\State $I_{ij} \gets \alpha \cdot I^{(r)}_{ij} + (1 - \alpha) \cdot I^{(c)}_{ij}$
\If{model\_type = "opt"}
\State $W^{\text{metric}}_{ij} \gets I_{ij} \cdot \varphi \cdot \tanh(|x|^\gamma) \cdot |x|^\beta$
\Else 
    \State $W^{\text{metric}}_{ij} \gets I_{ij} \cdot (\sqrt{x})^{\delta}$ 
\EndIf
\State \textbf{return} $W'$
\end{algorithmic}
\end{algorithm}

\begin{figure*}[t]
\centering
\includegraphics[width=1\textwidth]{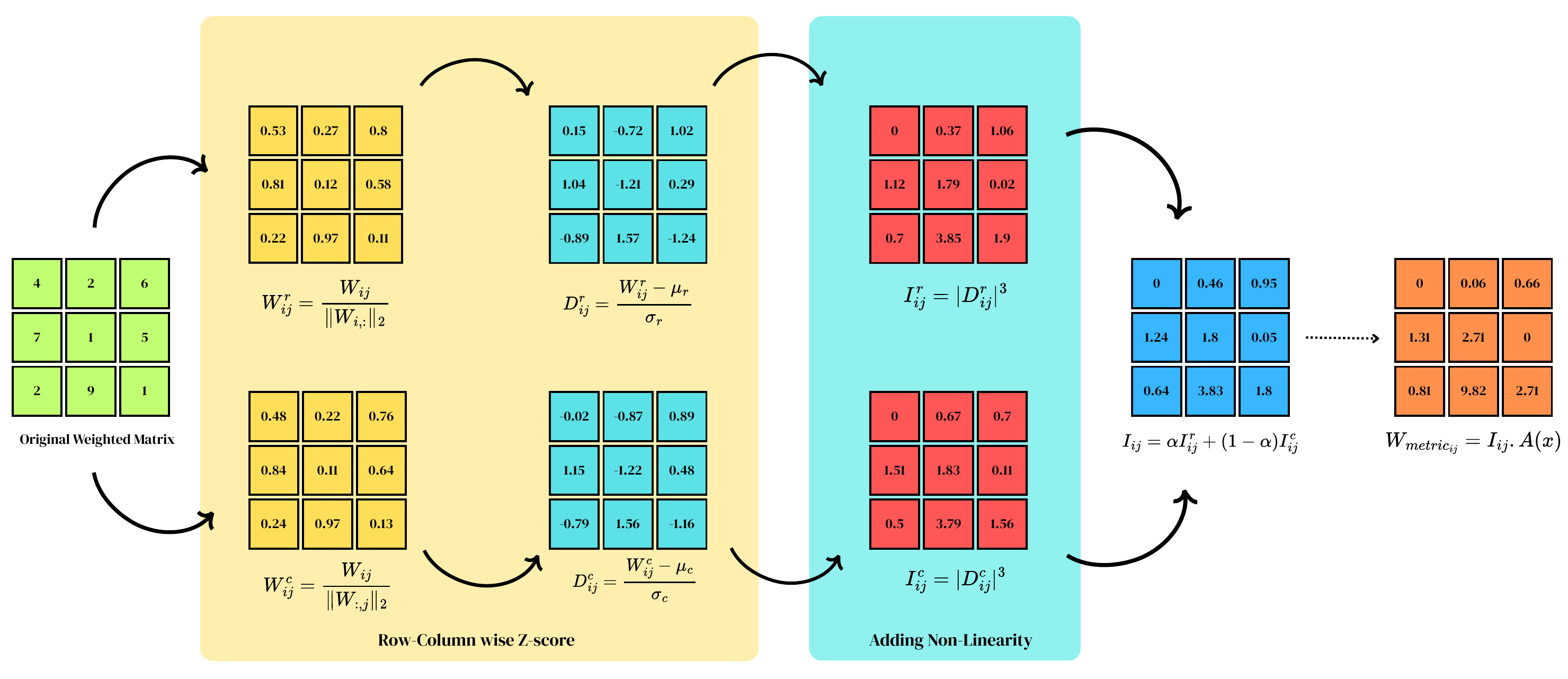}
\caption{A detailed example illustrating how the new important weight matrix is reconstructed from the original.}
\label{arc-design}
\end{figure*}

\subsection{Z-Score Pruning}
Z-Pruner begins by performing row-wise and column-wise L2 normalization on each weight matrix to stabilize their scales. Then, it computes a z-score for each weight based on its deviation from the mean, followed by a cubic amplification to highlight statistically significant outliers. To further align pruning with the model's dynamic behavior, Z-Pruner integrates activation scaling using two activation functions that better reflect the layer's actual contribution. An adaptive sparsity-aware balancing coefficient dynamically adjusts the pruning aggressiveness based on the target sparsity and layer type.

Z-Pruner significantly improves upon traditional magnitude pruning by incorporating a statistically-informed, context-aware framework for determining the importance of weights in large language models (LLMs). In contrast, magnitude pruning simply removes weights with the smallest absolute values; it operates under the assumption that these weights contribute the least to the model's output. This naive approach ignores the structural and functional context of each weight. It treats all weights equally, regardless of where they appear in the network or how they interact with others. This can lead to the pruning of weights that are small in magnitude but critical to the model's performance, especially in highly parameterized architectures where subtle interactions play a significant role. The detailed process of Z-Pruning is presented in Algorithm~\ref{algo1} and Figure~\ref{arc-design}.

\begin{algorithm}
\caption{Z-Pruner with masking Sparsity Ratio}
\label{algo2}
\textbf{Inputs:} weight matrix $W \in \mathbb{R}^{m \times n}$, importance metric $M \in \mathbb{R}^{m \times n}$, sparsity ratio $\rho \in [0,1]$, pruning mode (\texttt{per\_neuron}), reconstruction flag \\
\textbf{Output:} pruned weight matrix $W'$
\begin{algorithmic}[1]
\If{\texttt{per\_neuron} is True}
    \State Sort $M$ across each row
    \State Select smallest $\rho \cdot n$ indices per row
    \State Create binary mask $M_{\text{mask}}$
\Else
    \State Flatten $M$ and sort
    \State Threshold $\tau \gets$ $\rho$-quantile of $M$
    \State $M_{\text{mask}} \gets (M \leq \tau)$
\EndIf
\If{\texttt{reconstruction} is True}
    \State Apply reconstruction method with mask $M_{\text{mask}}$
\Else
    \State Set $W[M_{\text{mask}}] \gets 0$
\EndIf
\State \textbf{return} $W$
\end{algorithmic}
\end{algorithm}

Algorithm \ref{algo2} describes the pruning procedure used in Z-Pruner, which applies a sparsity-ratio-guided masking strategy. Given a weight matrix $W$ and its computed importance metric $M$, the algorithm supports two pruning modes: per-neuron or global. In per-neuron mode, the algorithm sorts the importance scores row-wise (corresponding to neuron-wise pruning) and identifies the lowest-scoring $\rho \cdot n$ weights in each row, constructing a binary mask that zeros these out. In the global mode, the importance scores are flattened, and a global threshold $\tau$ is determined using the $\rho$-quantile, masking out all weights below this threshold. An optional reconstruction flag allows a reconstruction method to be applied to the masked weights if desired; otherwise, the pruned weights are simply set to zero. The resulting masked weight matrix $W'$ is returned, achieving the specified sparsity in either neuron-local or global patterns while preserving the most important weights identified by the Z-score--based scoring scheme.

Initially, Z-Pruner uses the initial layer activation extractor to capture input embeddings and attention metadata from the first transformer layer using a calibration dataset, ensuring pruning reflects realistic input patterns and the first layer processes calibration inputs to compute activations for its sub-layers. In LLM models, layers are pruned iteratively. Starting with the first, using each layer's output as input for the next, ensuring consistent pruning across the model. The Z-pruner utilizes the extracted activation statistics from the previous layer and weight magnitudes combined to compute importance scores, and if the importance score of a logit is less than a threshold, the logits will be dumped.

Moreover, Z-Pruner introduces a dynamic sparsity-aware coefficient, which adjusts the balance between row-wise and column-wise importance based on the actual sparsity of the matrix. This adaptability is critical for preserving performance across different layers and model types. Finally, the use of model-specific activation functions to modulate the importance scores allows for further refinement, tailoring the pruning behavior to suit the nuances of architectures like OPT and LLaMA. This leads to a more intelligent and adaptive pruning process that prioritizes the preservation of essential weights, reducing performance loss and maintaining accuracy.

\subsection{Model Architectures}
Our experiments utilize a diverse set of transformer-based large language models (LLMs) from the LLaMA \cite{touvron2023llamaopenefficientfoundation} and OPT \cite{zhang2022optopenpretrainedtransformer} families, developed by Meta AI, to evaluate pruning techniques. The LLaMA models, ranging from 7 to 13 billion parameters, feature 32–40 attention heads, a vocabulary size of 32,000, and context lengths up to 8,192 tokens. They employ the Swish-Gated Linear Unit (SwiGLU) activation function and often incorporate grouped-query attention (GQA) for efficient handling of longer sequences, delivering state-of-the-art performance across benchmarks. In contrast, the OPT models, spanning 1.3 to 6.7 billion parameters, are designed as open-source alternatives with 16–32 attention heads, a vocabulary size of 50,257, and a context length of 2,048 tokens, utilizing the ReLU activation function for computational efficiency. The architectural diversity of LLaMA's advanced optimization and OPT's accessible design provides a robust framework for studying pruning impacts across varying model scales and configurations.

\section{EXPERIMENTS }

\subsection{Setup}
For the evaluation, we utilized cloud-based GPU setups. The OPT models were tested using a single T4 GPU, except OPT 6.7B. For OPT 6.7B, LLaMA-2 7B and LLaMA-3.1 8B, with dual T4 GPUs. Finally, the LLaMA-2 13B model was evaluated using an L40S GPU.

\noindent\textbf{Dataset:}
We used multiple datasets to evaluate the robustness of our algorithm. For perplexity evaluation, we used the WikiText-2\cite{merity2016pointersentinelmixturemodels} dataset across different families of LLaMA and OPT models. For model calibration, we employed the C4  \cite{raffel2023exploringlimitstransferlearning} dataset. To assess zero-shot performance, we evaluated on several benchmarks, including HellaSwag\cite{zellers2019hellaswagmachinereallyfinish}, BoolQ\cite{clark2019boolqexploringsurprisingdifficulty}, WinoGrande\cite{williams2018broadcoveragechallengecorpussentence}, MNLI\cite{zellers2019hellaswagmachinereallyfinish}, and WNLI\cite{nie2020adversarialnlinewbenchmark}.
The following is a description of the employed datasets: 

\begin{itemize}
    \item \textbf{WikiText-2:} A moderately sized language modeling dataset consisting of over 2 million tokens extracted from verified Wikipedia articles. It features well-formed, coherent English paragraphs without excessive markup or noisy data, making it a standard benchmark for evaluating language models on natural, high-quality text with long-range dependencies.
    
    \item \textbf{C4:} A large-scale, cleaned dataset derived from the Common Crawl web archive, totaling hundreds of gigabytes of English text. It filters out low-quality content, duplicates, and boilerplate, providing a diverse and high-volume source of real-world web data for pretraining and calibration of large language models.
    \item \textbf{HellaSwag:} A challenging dataset designed to test commonsense reasoning in natural language inference. It requires models to select the most plausible continuation of a given sentence from multiple choices, a task that is straightforward for humans but difficult for machines.  
    \item \textbf{BoolQ:} A dataset comprising naturally occurring yes/no questions paired with passages. Each example consists of a question, a passage, and the correct yes/no answer, aiming to evaluate a model's reading comprehension and inference capabilities.  
    \item \textbf{WinoGrande:} An expanded version of the Winograd Schema Challenge, containing 44,000 problems. It focuses on pronoun resolution tasks that require commonsense reasoning, designed to be more challenging and less susceptible to dataset-specific biases.  
    \item \textbf{MultiNLI:} A large-scale dataset with 433,000 sentence pairs annotated for textual entailment. It covers a range of genres, including fiction, government, and telephone speech, to test a model's ability to perform natural language inference across diverse contexts.   
    \item \textbf{WNLI:} A dataset derived from the Winograd Schema Challenge, part of the GLUE benchmark. It tests a model's ability to determine if one sentence entails another, focusing on pronoun resolution and requiring nuanced understanding of sentence structure.\\ 
    \end{itemize}
% \textbf{Implementation Details.}\\

\noindent\textbf{Evaluation:} To evaluate the performance of our proposed approach, we employ the \textbf{perplexity score} to measure the language modeling capability and \textbf{zero-shot accuracy} to assess generalization across different tasks based on a dataset. These metrics enable a comparative analysis across different families of large language models (LLMs). The evaluation metrics are described below:

\begin{itemize}
\item \textbf{Perplexity} quantifies how well a language model predicts a sequence of tokens. It is calculated as the exponentiation of the average negative log-likelihood of the predicted word probabilities. Specifically, for a sequence of tokens \( w_1, w_2, \dots, w_N \), the perplexity is defined as:

\[
PPL = \exp \left( - \frac{1}{N} \sum_{i=1}^{N} \log P(w_i | w_1, w_2, \dots, w_{i-1}) \right)
\]
where \( P(w_i | w_1, w_2, \dots, w_{i-1}) \) is the predicted probability of the \( i^{th} \) token given the previous tokens.

\item \textbf{Zero-Shot Accuracy} measures a model's ability to perform a task without any task-specific training examples. In a zero-shot setting, the model relies solely on its pre-trained knowledge and generalization capabilities to generate answers.
Given a task \( T \) with a set of input samples \( \{x_1, x_2, \dots, x_N\} \) and corresponding ground-truth labels \( \{y_1, y_2, \dots, y_N\} \), zero-shot accuracy is computed as:
\[
\text{Accuracy}_{\text{zero-shot}} = \frac{1}{N} \sum_{i=1}^{N} \mathbb{I}(f(x_i) = y_i)
\]
where \( f(x_i) \) is the model’s prediction for input \( x_i \), and \( \mathbb{I} \) is the indicator function that returns 1 if the prediction matches the true label, and 0 otherwise.
\end{itemize}

% In this study, we calculated the perplexity score on the \emph{WikiText-2} dataset, whereas zero-shot accuracy was computed across a variety of benchmark tasks, such as HellaSwag\cite{zellers2019hellaswagmachinereallyfinish}, BoolQ\cite{clark2019boolqexploringsurprisingdifficulty}, WinoGrande\cite{williams2018broadcoveragechallengecorpussentence}, MNLI\cite{zellers2019hellaswagmachinereallyfinish}, and WNLI\cite{nie2020adversarialnlinewbenchmark}.

\subsection{Results}

\begin{table*}[t]
\caption{Perplexity results on Wikitext2. We apply one-shot post-training pruning methods with 50\% unstructured sparsity on LLaMA-2, LLaMA3, and OPT models. \textcolor{red}{\textbf{Red bold}} texts denote the best (lowest) perplexity scores among all methods.}
\label{tab:performance}
\centering
\resizebox{\textwidth}{!}{%
\begin{tabular}{lcccccc}
\toprule
\textbf{Method}  & \textbf{OPT 1.3B} & \textbf{OPT 2.7B} & \textbf{OPT 6.7B}& \textbf{LLaMA-2 7B} & \textbf{LLaMA-2 13B} & \textbf{LLaMA-3.1 8B} \\
\midrule
Wanda\cite{sun2024simpleeffectivepruningapproach}  & 18.41 & 14.22 & 15.21 & 7.76 & 6.29 & 11.53 \\
SparseGPT\cite{frantar2023sparsegptmassivelanguagemodels} &  \textcolor{red}{\textbf{17.55}} & \textcolor{red}{\textbf{13.46}} & 11.62 & 7.01 & 6.03 & 9.86 \\
RIA\cite{zhang2024plugandplay} &  18.08 & 14.20 & 11.83 & 6.81 & 5.83 & 9.44 \\
Z-Pruner (Ours)  & 17.74 & 13.92 & \textcolor{red}{\textbf{11.60}}&  \textcolor{red}{\textbf{6.74}} & \textcolor{red}{\textbf{5.82}} & \textcolor{red}{\textbf{9.37}} \\
\bottomrule
\end{tabular}
}
\\[1ex]
% \centering \textit{Note*: Red denotes the best (lowest) perplexity scores among all methods.}
\end{table*}

% \begin{table*}[t]
% \caption{LLaMA-2-7B: Zero-Shot Performance of the model with unstructured 50\% sparsity compared to the dense model. Note*: Red denotes the best (highest) score among all the methods for each task.}
% \label{tab:zeroshot}
% \centering
% \footnotesize % Reduce font size
% \resizebox{\textwidth}{!}{%
% \begin{tabular}{lccccc|c}
% \toprule
% \textbf{Method} & \textbf{HellaSwag} & \textbf{BoolQ} & \textbf{WinoGrande} & \textbf{MNLI}  & \textbf{WNLI} & \textbf{Average} \\
% \midrule
% Magnitude\cite{han2015learningweightsconnectionsefficient} & 49.13 & 63.00 & 63.30 & 31.57 &  38.45 & 49.09 \\
% SparseGPT\cite{frantar2023sparsegptmassivelanguagemodels} & 52.75 & \textcolor{red}{\textbf{76.48}} & \textcolor{red}{\textbf{69.30}} & 38.57  & 40.85 & 55.59 \\
% Wanda\cite{sun2024simpleeffectivepruningapproach} & 50.32 & 75.05 & 67.80 & 38.14 &  42.25 & 54.71 \\
% RIA\cite{zhang2024plugandplay} & 52.04 & 74.22 & 68.27 & 39.31 & 42.25 & 55.22 \\
% Z-Pruner (Ours) & \textcolor{red}{\textbf{52.79}} & 74.98 & 68.51 & \textcolor{red}{\textbf{39.40}}& \textcolor{red}{\textbf{43.66}} & \textcolor{red}{\textbf{55.87}} \\
% \bottomrule
% \end{tabular}
% }
% \end{table*}

\begin{table*}[t]
\caption{LLaMA-2-7B: Zero-Shot Performance of the model with unstructured 50\% sparsity compared against prior approaches. \textcolor{red}{\textbf{Red bold}} texts denote the best (highest) score among all the methods for each task.}
\label{tab:zeroshot}
\centering
\footnotesize % Reduce font size
\resizebox{\textwidth}{!}{%
\begin{tabular}{lccccc|c|c}
\toprule
\textbf{Method} & \textbf{HellaSwag} & \textbf{BoolQ} & \textbf{WinoGrande} & \textbf{MNLI}  & \textbf{WNLI} & \textbf{Average} & \textbf{Pruning Time (min)} \\
\midrule
Magnitude\cite{han2015learningweightsconnectionsefficient} & 49.13 & 63.00 & 63.30 & 31.57 &  38.45 & 49.09 & \textcolor{red}{\textbf{4.51}} \\
SparseGPT\cite{frantar2023sparsegptmassivelanguagemodels} & 52.75 & \textcolor{red}{\textbf{76.48}} & \textcolor{red}{\textbf{69.30}} & 38.57  & 40.85 & 55.59 & 35.15 \\
Wanda\cite{sun2024simpleeffectivepruningapproach} & 50.32 & 75.05 & 67.80 & 38.14 &  42.25 & 54.71 & 13.47 \\
RIA\cite{zhang2024plugandplay} & 52.04 & 74.22 & 68.27 & 39.31 & 42.25 & 55.22 & 13.52 \\
Z-Pruner (Ours) & \textcolor{red}{\textbf{52.79}} & 74.98 & 68.51 & \textcolor{red}{\textbf{39.40}}& \textcolor{red}{\textbf{43.66}} & \textcolor{red}{\textbf{55.87}} & 11.81 \\
\bottomrule
\end{tabular}
}
\end{table*}

We conduct comprehensive experiments on our proposed algorithm, evaluating its performance across different families of large language models (LLMs), including LLaMA and OPT. Two primary evaluation metrics are used: perplexity and zero-shot accuracy. Perplexity is measured using the WikiText-2 dataset across six different LLMs. For zero-shot accuracy, we utilize the LLaMA-2-7B model and evaluate it on five benchmark datasets: HellaSwag, BoolQ, WinoGrande, MNLI, and WNLI.\\
\textbf{Perplexity Performance:}
Z-Pruner demonstrates strong performance on perplexity results across multiple models with 50\% unstructured sparsity. Notably, Z-Pruner achieves state-of-the-art  results with the lowest perplexity among all methods on OPT 6.7B (11.6) LLaMA-2 7B (6.74), LLaMA-2 13B (5.82), and LLaMA-3.1 8B (9.37). These results demonstrate the effectiveness of unstructured sparsity pruning in improving language model performance, particularly for larger models. All the models were evaluated and compared with other methods, including Z-Pruner, as shown in Table \textcolor{blue}{\ref{tab:performance}} \\
\textbf{Zero-shot Performance:} Table \textcolor{blue}{\ref{tab:zeroshot}} presents the zero-shot performance of the LLaMA-2-7B model under 50\% unstructured sparsity. The evaluation spans five benchmark datasets: HellaSwag, BoolQ, WinoGrande, MNLI, and WNLI. The final column of the table reports the average performance across these datasets. Notably, Z-Pruner achieves the highest performance on 3 out of the 5 datasets and also records the best average performance across all datasets, underscoring its effectiveness in zero-shot settings. For zero-shot performance on LLaMA-2-7B under 50\% sparsity, Z-Pruner sets new state-of-the-art accuracy on HellaSwag (52.79), MNLI (39.40), WNLI (43.66), and achieves the highest overall average performance (55.87) compared to all other methods.\\
\textbf{Comparision with Sparsegpt:} While our approach outperforms SparseGPT in terms of overall performance, particularly on benchmarks such as OPT-6.7B, LLaMA-2-7B/13B, and LLaMA-3.1-8B with respect to perplexity—SparseGPT shows better results on smaller models like OPT-1.3B and 2.7B. Additionally, in zero-shot settings, SparseGPT achieves higher accuracy than Z-Pruner on tasks such as BoolQ and WinoGrande using LLaMA-2-7B. However, SparseGPT is computationally intensive, requiring significant weight updates, which leads to longer runtimes, as shown in our runtime analysis section. This makes it less practical for real-world applications. In contrast, Z-Pruner scales better with model size, delivering stronger performance on larger models, comparable or better results on smaller models, and significantly reduced inference time, making it more suitable for practical deployment.

\noindent\textbf{Running Time Analysis:}
To evaluate the efficiency of each pruning algorithm, we report the actual runtime measured on one of the large models used in our experiments, LLaMA-2-7B. Each method was tested using 166 calibration samples. As shown in Table \ref{tab:zeroshot}, SparseGPT had the longest pruning time, taking 35.15 minutes. In contrast, Wanda and RIA achieved significantly faster runtimes of 13.47 and 13.52 minutes, respectively. Z-Pruner outperformed all other advanced methods with the fastest pruning time of 11.81 minutes. This relatively low runtime suggests that Z-Pruner is less memory-intensive, contributing to its overall efficiency. Although magnitude pruning achieved the shortest pruning time of just 4.51 minutes, its simplistic approach does not capture the actual relationship between weights which resulted in the lowest average accuracy on zero-shot tasks. While SparseGPT may offer strong performance in certain scenarios, its high memory consumption leads to substantially longer pruning durations.

\begin{figure}
\centering
\includegraphics[width=0.48\textwidth]{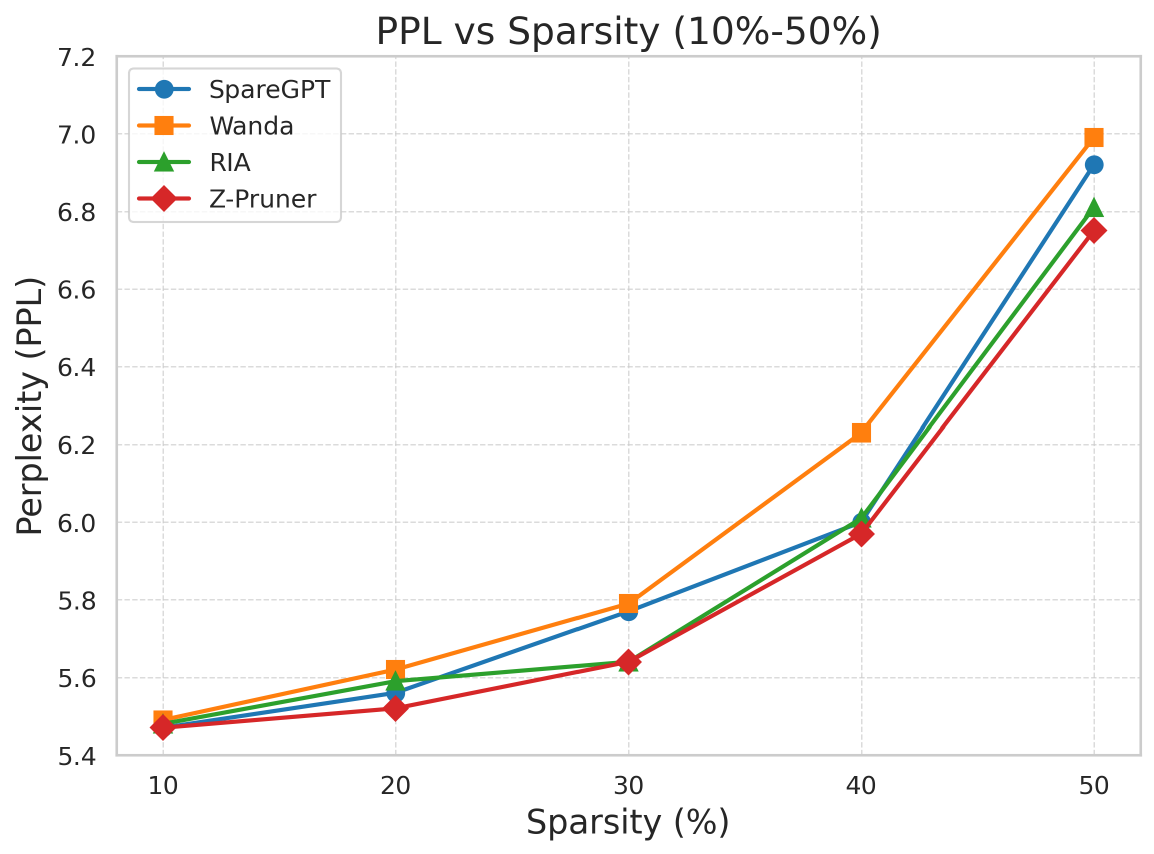} % Adjust the width as needed
\caption{Perplexity performance comparison for LLaMA-2-7B under different Sparsity ratio (10\%–50\%).}
\label{pplvssparsity}
\end{figure}

\noindent\textbf{Sparsity-Robustness Analysis:}
To assess the robustness of pruning algorithms under moderate sparsity, we evaluated the perplexity (PPL) behavior of four techniques - SpareGPT, Wanda, RIA, and Z-Pruner, across different sparsity levels ranging from 10\% to 50\% in LLaMA-2-7B. The results, depicted in Figure \textcolor{blue}{ \ref{pplvssparsity}}, reveal that all methods maintain high fidelity in language modeling performance, with perplexity exhibiting only marginal increases across this range. Importantly, Z-Prune consistently achieves the lowest perplexity values at each sparsity level, thereby demonstrating superior resilience to parameter ablation. RIA and SpareGPT also perform competitively, exhibiting slightly elevated PPL relative to Z-Pruner, yet maintaining performance within a narrow margin. In contrast, Wanda, although stable at lower sparsity, displays a comparatively larger perplexity increment as sparsity approaches 50\%, indicating a heightened sensitivity to aggressive pruning.

Notably, across the 10\%-50\% sparsity regime, the observed perplexity variations remain bounded within a tight band (approximately 5.4 to 7.0), underscoring the efficacy of modern pruning techniques in preserving model quality even under substantial sparsification. The minimal perplexity degradation observed suggests that up to 50\% of model weights can be pruned without materially impacting predictive performance, thus validating the feasibility of structured sparsity as a compression strategy. These findings have significant implications for the deployment of large language models in latency- and memory-constrained environments, where maintaining accuracy while reducing computational overhead is critical.

\subsection{Ablation Studies}

\begin{figure}[!t]
\centering
\includegraphics[width=0.5\textwidth]{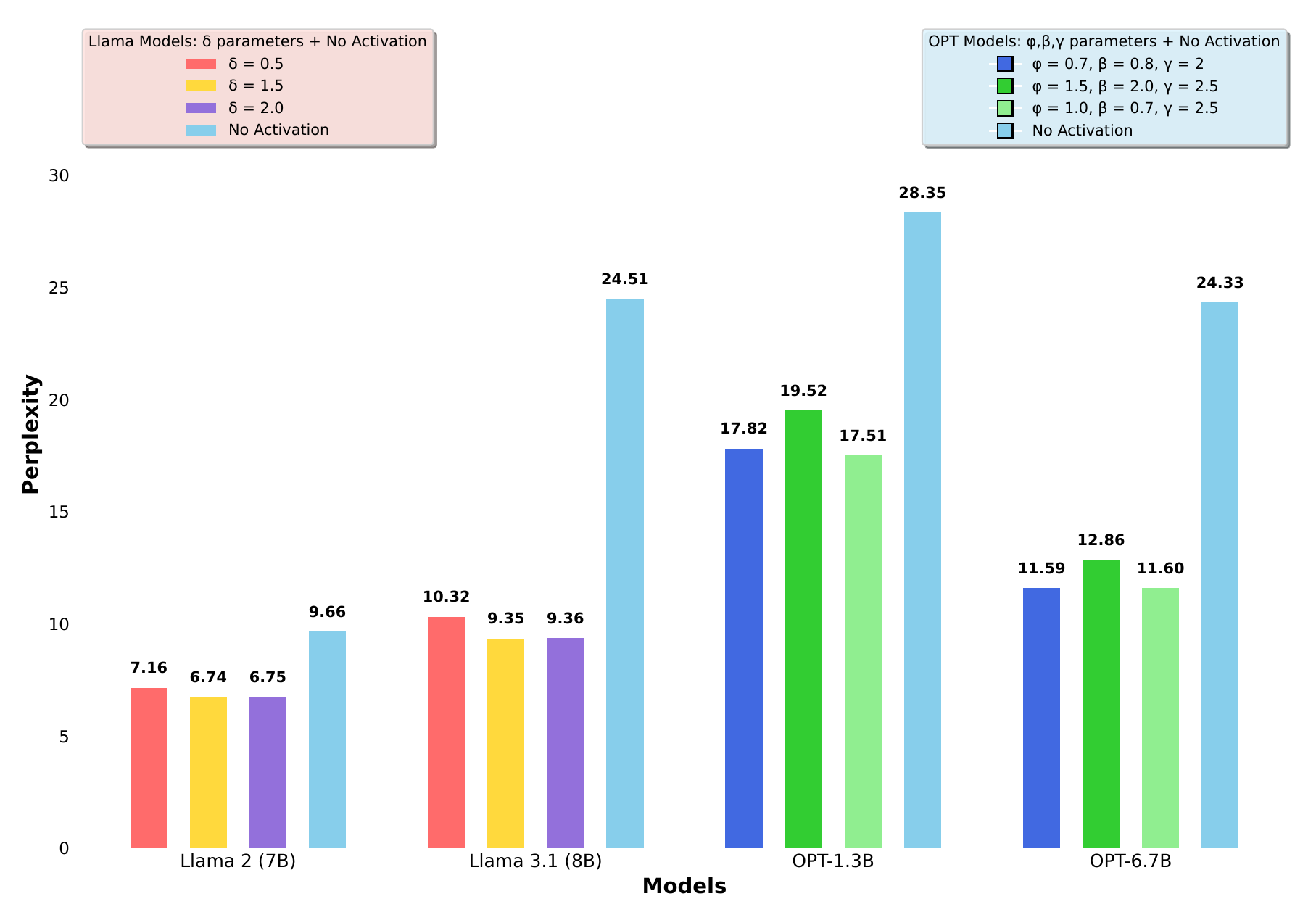} % Adjust the width as needed
\caption{Ablation Studies of Z-Pruner on LLaMA-2-7B, LLaMA-3.1-8B, OPT-1.3B, and OPT-6.7B models.}
\label{ablation}
\end{figure}

In this ablation study Figure \textcolor{blue}{\ref{ablation}}, we tested different activation and scaling parameters to identify optimal configurations for both LLaMA and OPT models. For LLaMA, we explored $\delta$ values of 0.5, 1.5, and 2.0 after preliminary tests showed these ranges provided stable gradient behavior and improved scaling of activations, ultimately finding $\delta=1.5$ delivered the best trade-off, reducing perplexity to 9.35 for LLaMA 3.1-8B and 6.74 for LLaMA-2-7B. For OPT models, we evaluated combinations of $\varphi$, $\beta$, and $\gamma$ based on prior activation scaling literature, testing values such as $ \varphi=0.7$, $\beta=0.8$, $\gamma=2$ and higher variants to capture nonlinearities better, with the best results achieved at $ \varphi=1.0$, $\beta=0.7$, $\gamma=2.5$, which brought perplexity down to 17.51 for OPT-1.3B and 11.60 for OPT-6.7B. We included a no-activation baseline for fair comparison, showing that adding these activations improved performance significantly across all models, confirming that thoughtful parameter selection and scaling greatly enhance post-training calibration.

\subsection{Discussion}
This study presents Z-Pruner, a novel unstructured pruning approach tailored for large language models (LLMs), with extensive evaluation across multiple model families and benchmarks. Our results show that Z-Pruner consistently outperforms existing methods in both perplexity and zero-shot accuracy, particularly under high sparsity conditions. When evaluated on perplexity using the WikiText-2 dataset, Z-Pruner achieves state-of-the-art results across large-scale models, including OPT-6.7B, LLaMA-2-7B/13B, and LLaMA-3.1-8B, demonstrating its robustness and effectiveness in maintaining quality under significant parameter reduction.
In zero-shot settings, Z-Pruner also achieves top performance on several benchmarks using LLaMA-2-7B, outperforming other pruning methods on HellaSwag, MNLI, and WNLI, and achieving the highest average accuracy across five datasets. While SparseGPT performs better on selective tasks and smaller models, it incurs significantly higher computational costs due to its memory-intensive pruning process. In contrast, Z-Pruner delivers competitive or superior performance with much faster runtimes. These results underscore Z-Pruner’s potential as a practical and scalable pruning solution for deploying LLMs in resource-constrained environments without compromising model quality.

\section{Limitations}
The current approach, while promising, comes with several limitations. Firstly, the hyperparameters used in our method have not been rigorously tested or optimized. A more thorough tuning process could significantly enhance performance. Secondly, the pruning formula itself is not yet fully generalized; we are still actively experimenting with different activation functions to identify the most robust and adaptable formulation. Thirdly, the approach struggles with newer models like LLaMA 3.2  due to architectural differences in their layer mechanisms, which likely require layer-wise error correction. Although the method still offers minimal effectiveness according to external evaluations, its compatibility is limited. Lastly, to truly establish its utility, the method must be developed further to consistently achieve state-of-the-art results across diverse benchmarks and model architectures.

\section{Conclusion}
In this work, we presented Z-Pruner, a novel post-training pruning method that enables efficient sparsification of large language models without the need for retraining. By leveraging a statistically-informed approach using Z-scores normalized across both rows and columns of weight matrices, Z-Pruner effectively preserves important parameters while eliminating redundancies. Our experiments demonstrate that Z-Pruner consistently outperforms traditional magnitude pruning and state-of-the-art pruning methods like SparseGPT and Wanda across multiple benchmarks and model sizes, particularly excelling in perplexity and zero-shot downstream task performance under unstructured 50\% sparsity. Moreover, Z-Pruner achieves these results with faster pruning times, making it highly suitable for practical use case. While promising, the method has limitations in terms of hyperparameter generalization and compatibility with newer model architectures. Future work will focus on optimizing hyperparameters, expanding support for heuristic approach to achieve better performance, and enhancing adaptability across a broader range of LLMs. Overall, Z-Pruner offers a powerful and efficient solution for compressing LLMs, balancing resource savings with strong task performance.

\bibliographystyle{abbrv}   
{\tiny
\bibliography{ref}
}
\vfill
\end{document}